\documentclass{article}
\usepackage{spconf, amsmath, graphicx}
\usepackage{xcolor, mathrsfs}
\usepackage{import, subfigure}
\usepackage{amssymb, subcaption}
\usepackage{inputenc, soul}
\usepackage{multirow, comment}
\usepackage{booktabs, url}
\usepackage{balance}
\usepackage[T1]{fontenc}

\title{Active Listener: Continuous Generation of Listener's Head Motion Response in Dyadic Interactions}
\name{Bishal Ghosh, Emma Li, Tanaya Guha}
\address{University of Glasgow, UK}

\begin{document}
\ninept
\maketitle

\begin{abstract}
A key component of dyadic spoken interactions is the contextually relevant non-verbal gestures, such as head movements that reflect a listener's response to the interlocutor's speech. Although significant progress has been made in the context of generating co-speech gestures, generating listener's response has remained a challenge. We introduce the task of generating continuous head motion response of a listener in response to the speaker's speech in real time. To this end, we propose a graph-based end-to-end crossmodal model that takes interlocutor's speech audio as input and directly generates head pose angles (roll, pitch, yaw) of the listener in real time. Different from previous work, our approach is completely data-driven, does not require manual annotations or oversimplify head motion to merely nods and shakes. Extensive evaluation on the dyadic interaction sessions on the IEMOCAP dataset shows that our model produces a low overall error (4.5 degrees) and a high frame rate, thereby indicating its deployability in real-world human-robot interaction systems. Our code is available at \url{https://github.com/bigzen/Active-Listener}

\end{abstract}
\section{Introduction}
\label{sec:intro}

\begin{figure}[t]
    \centering \includegraphics[width=0.8\linewidth, trim ={5cm 3.7cm 6cm 4cm}, clip=true]{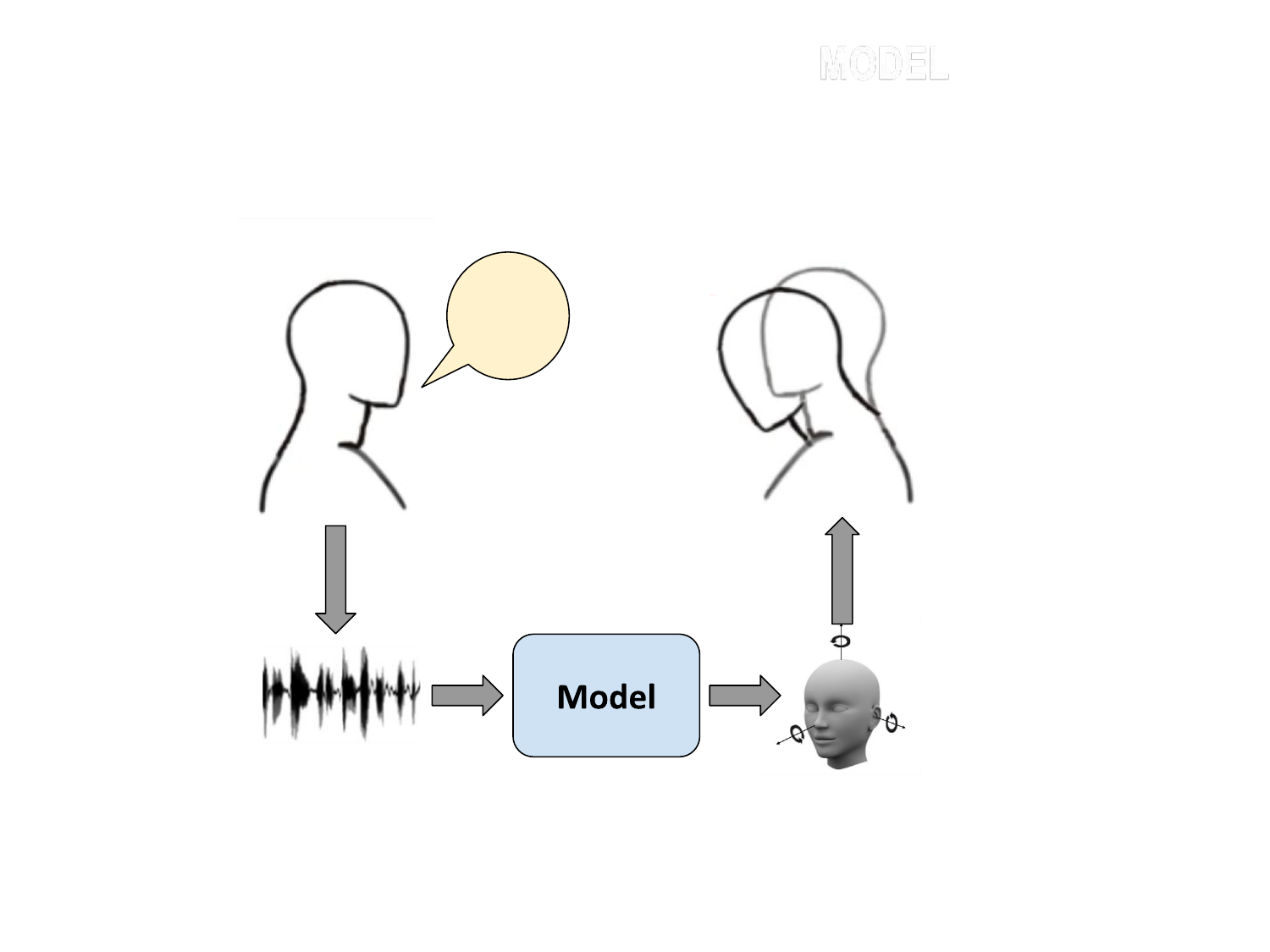}
    \vspace{-1mm}
    \caption{We introduce the task of generating continuous head motion response of a listener solely based on speaker's speech in a dyadic interaction. Different from past work, we present a completely data-driven approach that generates 3D head pose sequence in real time.}
    \label{fig:int}
\end{figure}

An important component of dyadic interactions (human-human or human-agent) is the contextually relevant verbal and non-verbal cues that reflect a listener's response (aka \emph{backchannels}) to the interlocutor's speech. Head movements, such as nods/shakes are common backchannel gestures that a listener uses to provide feedback and to maintain the flow of communication while demonstrating active engagement. Although significant progress has been made in the context of generating speaker's head and body movements (co-speech gestures) \cite{belpaeme2024}, generating appropriate backchannel gestures has remained a challenge. The \textbf{goal} of this work is to \emph{generate continuous head motion response of a listener in response to the speaker's speech} in real time (see Fig.~\ref{fig:int}). 
\par
The majority of existing work on listener's head gesture understanding is formulated as a binary task i.e., to predict the presence or absence of backchannel head gestures using speaker's speech and gestures as inputs \cite{belpaeme2024,zhou2022responsive,44}. If head gesture is present, the next step usually involves predicting the type of head gesture (e.g., nods/shakes). For example, a semi-supervised approach using both speech and speaker's facial gestures is proposed to first predict backchannel opportunity i.e., when a head gesture response should be produced, and next identify which backchannel gesture should be made such as nods or shakes or other cues \cite{44}. Eye gaze (manually annotated) has been shown as an useful non-verbal cue to predict head nods using a sequential probabilistic model \cite{46}. Works have also used multitask learning to jointly predict backchannel head gestures, vocalized fillers and turn-taking \cite{42,43}. However, all the work above considers only a binary detection of of head gestures (presence vs. absence; nod vs shake). This reflects (i) a `lumping approach' \cite{49} that reduces the head gesture variability to two simple patterns, and (ii) produces sporadic head gestures that is not natural or human-like. 
\begin{figure*}[!t]
    \centering
    \includegraphics[width=0.95\linewidth]{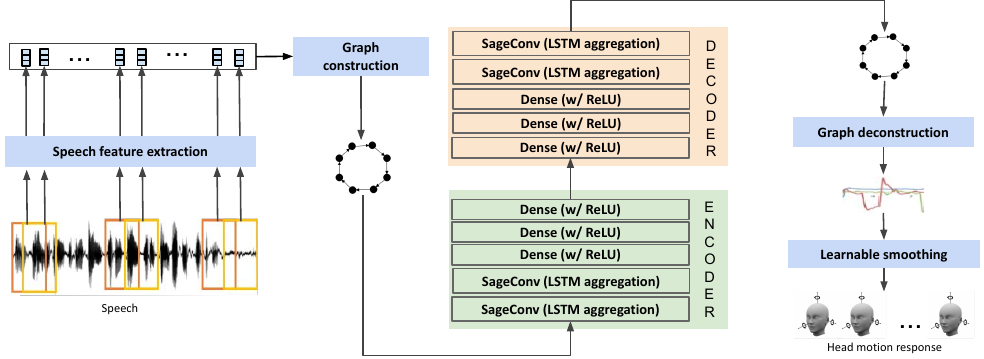}
    \caption{Overview of our model to generate listener's head motion response from speaker's speech. Speech is represented as a cycle graph which uses a GNN-based encoder-decoder architecture to generate head motion in terms of a time series of head pose angles. The graph architecture produces a compact yet accurate model facilitating real time generation.}
    \label{fig:model}
\end{figure*}
\par Considering the broader field of backchannel generation (including but not limited to head gestures), we note that early research focused on rule-based approaches \cite{belpaeme2024}, while the modern trend is to adopt data-driven approaches. The existing data-driven approaches, however, rely heavily on manual annotations i.e., labeling presence or absence of nods/shakes \cite{belpaeme2024}. This creates a bottleneck as suitably annotated data is unavailable due to the labour-intensive nature of the annotation task. Recently, continuous backchannel generation has been addressed through 3D mesh generation \cite{zhou2022responsive}, where the authors focused on creating a dataset (called ViCo) for listener-centric head motion generation. Another recent work proposed a unified modeling framework based on vector quantized variational autoencoder (VQ-VAE) to generate listener's behaviour in videos \cite{tran2024dyadic}. Both of these work on continuous listener's response generation focused on online interactions, and use speakers videos containing speech and facial gestures. In contrast, our work focuses on in-person dyadic interactions and uses only speaker's speech as input.
\par In this work, we propose a graph-based end-to-end model for generating continuous head motion response of a listener using the speaker's speech as input. Our crossmodal model follows an encoder-decoder architecture that directly generates head pose angles (roll, pitch, yaw) of the listener at a minimum rate of 86 frames per second (real time). Our approach overcomes the limitations of past work as it does not require any manual annotation or reduce head motion to only nods/shakes. To the best of our knowledge, this is the first work on continuous, real time listener's head response generation using speaker's speech alone. For validation, we use the IEMOCAP dataset, which contains speech and motion capture recordings of in-person dyadic interactions. Our contributions are:
\vspace{-2mm}
\begin{itemize}
    \item A graph-based end-to-end model to generate \textit{continuous} head motion response of a listener during dyadic interactions.
    \vspace{-1mm}
    \item An efficient model that runs real time and generates head pose sequence with an average absolute error of $4.5$ degrees. 
    \vspace{-1mm}
    \item Superior performance for generalised (speaker independent) model using only speaker's speech as input  without the need of any annotations.
\end{itemize}
\vspace{-2mm}
\section{Proposed Approach}
\label{sec:format}
\begin{table*}[ht!]
    \caption{Performance of our proposed generalised (subject-independent) head motion response generation model for different feature sets. The results are presented in terms of MAE (mean $\pm$standard deviation) across 5-fold cross validation. The speed indicates end-to-end generation speed in terms of fps.} \vspace{-1mm}
\label{tab:results_main}
 \renewcommand*{\arraystretch}{1.2}
\centering
\begin{tabular}{l c | c c c|c| c c}
\toprule
 {\bf Model}  & {\bf Features}  & {\bf Roll} ($\phi$) & {\bf Pitch} ($\psi$) & {\bf Yaw} ($\theta$) & \bf All & \bf Params & {\bf Speed} \\ 
\midrule
Linear {\footnotesize(baseline)} & \multirow{5}{*}{MFCC} & $23.40\pm16.01$ & $23.94\pm16.50$ & $23.08\pm15.80$ & $23.47\pm16.10$ &- &-\\
LSTM {\footnotesize(baseline)} & & $8.69\pm4.62$ & $8.30\pm4.78$ & $7.95\pm4.77$ & $8.31\pm4.73$& $1.0$M & $14,107$\\
Ours {\footnotesize(w/o smooth)} & & $\mathbf{6.49}\pm4.23$ & $6.37\pm4.57$ & $6.20\pm4.56$ &  $6.35\pm4.45$ &$0.8$M & $18,490$\\
Ours {\footnotesize(w/o cos sim)} & & $7.15\pm2.68$&$7.15\pm3.38$&$6.81\pm3.42$& $7.04\pm3.16$& $0.8$M & $18,490$\\
\bf Ours & & ${6.64\pm4.29}$ & $\mathbf{6.08}\pm4.55$ & $\mathbf{6.03}\pm4.64$ & $\mathbf{6.25}\pm4.49$ & $0.8$M & $18,490$\\
\midrule
Linear & \multirow{5}{*}{eGEMAPS$^*$} & $11.80\pm6.97$ & $12.21\pm7.59$ & $11.61\pm7.02$ & $11.87\pm7.19$&- &-\\
LSTM  & & $7.90\pm5.03$ & $7.12\pm5.75$ & $7.16\pm5.47$ & $7.39\pm5.42$& $1.0$M & $91$\\
Ours {\footnotesize(w/o smooth)} & & $\mathbf{6.96}\pm2.21$ & $\mathbf{6.51}\pm3.05$ & $\mathbf{6.34}\pm2.57$ & $\mathbf{6.60}\pm2.61$ & $0.9$M & $86$\\
Ours {\footnotesize(w/o cos sim)} & & $7.55\pm3.44$&$7.04\pm3.58$&$7.03\pm3.42$ & $7.21\pm3.48$& $0.9$M & $86$\\
\bf Ours & & $8.28\pm1.62$ & $6.89\pm2.11$ & ${9.40\pm1.94}$ & ${8.19\pm1.89}$& $0.9$M & $86$\\
\midrule
Linear &\multirow{5}{*}{Wav2vec2}& $32.35\pm23.52$ & $34.57\pm25.10$ & $30.91\pm22.63$ & $32.61\pm23.75$&- &-\\
LSTM &  & $9.49\pm3.96$ & $8.07\pm4.01$ & $8.83\pm4.11$ & $8.80\pm4.03$& $1.5$M & $1,354$\\
Ours {\footnotesize(w/o smooth)} & & $3.77\pm4.01$&$4.62\pm3.48$&$6.80\pm6.09$& $5.06\pm4.53$ & $3.3$M & $1,424$ \\
Ours {\footnotesize(w/o cos sim)} & & $\mathbf{3.40}\pm3.60$ & $\mathbf{3.93}\pm2.55$ & $\mathbf{6.13}\pm5.87$ & $\mathbf{4.49}\pm4.01$& $3.3$M & $1,424$\\
\bf Ours & & $3.41\pm3.58$ & $4.00\pm2.58$ & ${6.24}\pm5.82$ & $4.55\pm3.99$& $3.3$M & $1,424$\\
\bottomrule
     $^*${\footnotesize uses CPU implementation}
\end{tabular}

\end{table*}
\vspace{-2mm}
Our model follows a cross-modal encoder-decoder paradigm to generating continuous head motion response of a listener in a dyadic interaction. Our model is based on a graph architecture combined with an LSTM aggregation (see Fig.~\ref{fig:model}). We adopt a graph-based approach as they have been shown to yield high performance with fewer parameters in various speech-based applications \cite{shirian2022self,shirian2021compact}. The encoder takes a short segment of speaker's speech as input in the form of a line graph \cite{shirian2021compact}, while the decoder generates the corresponding head pose (pitch, yaw and roll) of the listener for that speech segment. Below, we describe each component in detail.
\vspace{-3mm}
\subsection{Graph construction}
\vspace{-1mm}
Following past work \cite{shirian2021compact}, we convert the input \textbf{speech} signal to a \emph{directed cycle graph} $\mathcal{G}_{s}=(\mathcal{V,E})$ through a frame-to-node transformation. Thus the set $\mathcal{V}=\{v_{s_i}\}_{i=1:M}$ contains $M$ nodes, where each node  corresponds to a frame (small, overlapping segment) of the speech signal, and the set $\mathcal{E}$ contains all edges between the nodes, each node having a edge directed from its previous node to itself with weight 1. This unidirectional graph is used to enable a real time modeling as only information from the past is assumed to be available. The adjacency matrix of $\mathcal{G}_s$ is denoted by $\mathbf{A}_s\in\mathbb{R}^{M\times M}$ (see eq.1) where an element $(\mathbf{A}_s)_{ij}$ denotes the edge weight connecting $v_{s_i}$ and $v_{s_j}$. Each node $v_{s_i}$ is associated with a \emph{node feature vector} $\mathbf{v}_{s_i}\in \mathbb{R}^m$, which contains embeddings extracted from the corresponding speech frame (details in Section \ref{sec:experiments}).\\
\vspace{-2mm}
\begin{equation}
        \mathbf{A}_s=
        \begin{bmatrix}
        0&1&0&\cdots&0\\
        0&0&1&\cdots&0\\
        0&0&0&\cdots&0\\
        \vdots&\vdots&\vdots&\ddots&\vdots\\
        1&0&\cdots&0&0\\
        \end{bmatrix}
    \end{equation}
\par
Similarly, the \textbf{head motion} is converted to a cycle graph $\mathcal{G}_h$ with $N$ nodes $\{v_{h_i}\}_{i=1:N}$ and directed edges as before. Here, each node corresponds to a time step in the head motion signal. The node feature vector associated with each node $\mathbf{v}_{h_i}=[\phi_i, \psi_i,\theta_i]^T$ contains the Euler angles (roll $\phi$, pitch $\psi$ and yaw $\theta$) used to represent head pose. Note that we resample the head motion signal so as to have $M=N$ (details in Section \ref{sec:experiments}). Therefore the $\mathbf{A}_h=\mathbf{A}_s$, where $\mathbf{A}_h$ is the adjacency matrix for $\mathcal{G}_h$.
\vspace{-3mm}
\subsection{Encoder-Decoder with learnable smoothing} 
\vspace{-1mm}
The \textbf{encoder} takes $\mathcal{G}_s$ as input and uses the SAGEConv architecture \cite{sageconv} as its backbone. We use two SAGEConv layers which involve graph convolution with LSTM aggregation \cite{sageconv}. This is followed by three dense layers with ReLU activation. We stack the node features from the second SAGEConv layer to feed to the dense layers. The encoder thus yields a representation $\mathbf{Z}\in \mathbb{R}^{M\times P}$, where $P$ is the number of neurons in the final dense layer. The \textbf{decoder} takes $\mathbf{Z}$ as input and learns to estimate $\mathcal{G}_h$. It is designed to mirror the encoder with three dense layers with ReLU, followed by two SAGEConv layers. We deconstruct the output of the dense layers to finally reconstruct $\mathcal{G}_h$ with $N(=M)$ nodes, where the $i^{th}$ node is associated with node vector $\hat{\mathbf{v}}_{h_i}=[\hat{\phi}_i, \hat{\psi}_i,\hat{\theta}_i]^T$ representing the head pose angles at every time step (corresponds to each node).

\par Next, we smooth the head motion output
$\{\hat{\mathbf{v}}_{h_1}, \hat{\mathbf{v}}_{h_2}\cdots \hat{\mathbf{v}}_{h_N}\}$, where $\hat{\mathbf{v}}_{h_i}\in\mathbb{R}^3$ using a learnable Gaussian kernel. Three 1D Gaussian kernels with 0 mean and variances $\sigma_r, \sigma_p, \sigma_y$ are convolved along each dimension of $\hat{\mathbf{v}}_{h_i}$ through 1D convolution, where $\sigma_r, \sigma_p, \sigma_y$ are learnable parameters. Let's denote the smoothed head motion time series as $\{\tilde{\mathbf{v}}_{h_i}\}_{i=1:N}$.
\vspace{-3mm}
\subsection{Loss function} 
\vspace{-1mm}
To train the model above, we minimize following loss function combining mean squared error (MSE) and cosine similarity:\\
\vspace{-1mm}
\begin{align}
    L&= MSE(\mathbf{v}_{h_i}, \tilde{\mathbf{v}}_{h_i}) + (1-L_{sim})\\
    L_{sim}&=\sum_i s([\phi_i, \psi_i, \theta_i],[\hat{\phi}_i,\hat{\psi}_i,\hat{\theta}_i])
\end{align}
where $s(.)$ denotes cosine similarity.
\section{Experiments}
\label{sec:pagestyle}

\label{sec:experiments}
\vspace{-1mm}
\subsection{Dataset}
\vspace{-1mm}
We repurposed the IEMOCAP \cite{iemocap} dataset, originally developed for speech emotion analysis, for our task. This dataset contains five sessions of dyadic interactions recorded in form of audio, visual and motion capture (MoCap) data. The interaction sessions are acted by two professional actors using a script, but also has spontaneous moments. The interactions are annotated with arousal/valence/dominance value, categorical labels for emotions, MoCap data for one participant, transcriptions and segmented audio files for the conversation. Our work focuses on the listener's head motions so we filtered out the dataset where the MoCap data was available for the listener. We achieved this by selecting the data where speaker code did not match with the MoCap participant code for the output and selecting segmented audio file of that data for input. This selection yielded 4072 speech-MoCap pairs.
\vspace{-2mm}
\subsection{Features}
\vspace{-1mm}
\textbf{Speech:} Our graph-based architecture described above is tied to any particular feature. We experiment with the following speech features to study the effectiveness of our model across different features and the performance of different features: \\
$\bullet$ \textbf{MFCC.} We extracted 28 mel coefficients using a window size of 64ms and hop length 33ms using Torchaudio \cite{torchaudio}.\\
$\bullet$ \textbf{eGEMAPS.} This feature set contains 88 features following the Geneva Minimalistic Acoustic Parameter Set \cite{egemaps} computed on audio segment of 64ms at 33ms interval.\\
$\bullet$ \textbf{Wav2Vec2.} We used the pretrained Wav2Vec2 \cite{wav2vec} model to extracted a 512-dimensional feature vector corresponding to each speech segment (node).\\

\vspace{-2mm}
\noindent\textbf{Head motion:} For head motion, we use roll, pitch and yaw head angles provided in the dataset as MoCap recording. The original sampling rate of 120Hz was resampled at 50Hz when using Wav2vec2 features and at 30Hz otherwise. The variable sampling rate arose due to our inability to change the output sample rate of pretrained Wav2Vec2 model. 
\vspace{-2mm}
\subsection{Baselines}
\vspace{-1mm}
As this is the first work on continuous generation of listener's head motion from speaker's speech, we develop two baselines to compare with our proposed model.\\

\vspace{-3mm}
\noindent\textbf{Linear regression:} The linear regression model is one of the simplest models available, thus we used it to form our first baseline. With $M$ features we learnt $M+1$ weights. Due to size of the data, we used a multi-step optimisation approach with Adam optimizer. \\

\vspace{-3mm}
\noindent\textbf{LSTM baseline:} The LSTM baseline model consists of an encoder and a decoder. The encoder consists of a single unidirectional LSTM layer (256 nodes) followed by 2 fully connected dense layers (with 384 and 128 nodes). The decoder contains 2 dense layers (with 128 and 6 nodes) followed by a unidirectional LSTM layer with three hidden nodes. ReLU activation and batch normalization are applied before every hidden layer. Smoothing is also applied to the output using 1D convolution with a learnable Gaussian kernel as in our proposed model. This model uses the same loss function as our proposed model.
\vspace{-3mm}
\subsection{Implementation details}
\vspace{-1mm}
\textbf{Parameter settings:} Our model's encoder consists of 2 graph convolution layers (128 and 256 nodes) followed by 3 fully connected dense layers (384, 128 and 128 nodes). The decoder contains 3 dense layers (128, 384 and 128 nodes) followed by 2 graph convolution layer (256 and 3 nodes). We use Adam optimizer with a learning rate of $10^{-4}$, a reduce learning rate on plateau learning rate scheduler with reduction factor 0.5, patience 50, relative threshold 0.01. Training was performed with batch size of 64 and training cutoff at 500 epochs or epoch when learning rate drops below $10^{-6}$ whichever is smaller.\\

\vspace{-2mm}
\noindent\textbf{Training protocol:} We use a subject-independent setting that can achieve higher generalizability. 
We use a five-fold cross validation with four sessions of the dataset used for training and the remaining one for testing. There is no subject overlap between the training and the test set. We report average error across the five folds.
\\

\vspace{-2mm}
\noindent\textbf{Evaluation metric:} We use mean absolute error (MAE) between the predicted head pose angles and the ground truth to evaluate the model performance. This metric is adopted for higher interpretability, where an error $x$ indicates that the prediction is off by $x$ degrees. The 5 fold cross validation was performed five times iterating over the sessions to form mutually exclusive test sets and the reported results were averaged over all the five test sets.
%
\begin{figure}[tb]
        \centering 
        \includegraphics[width=\linewidth, trim ={2mm 3mm 3mm 2mm}, clip=true]{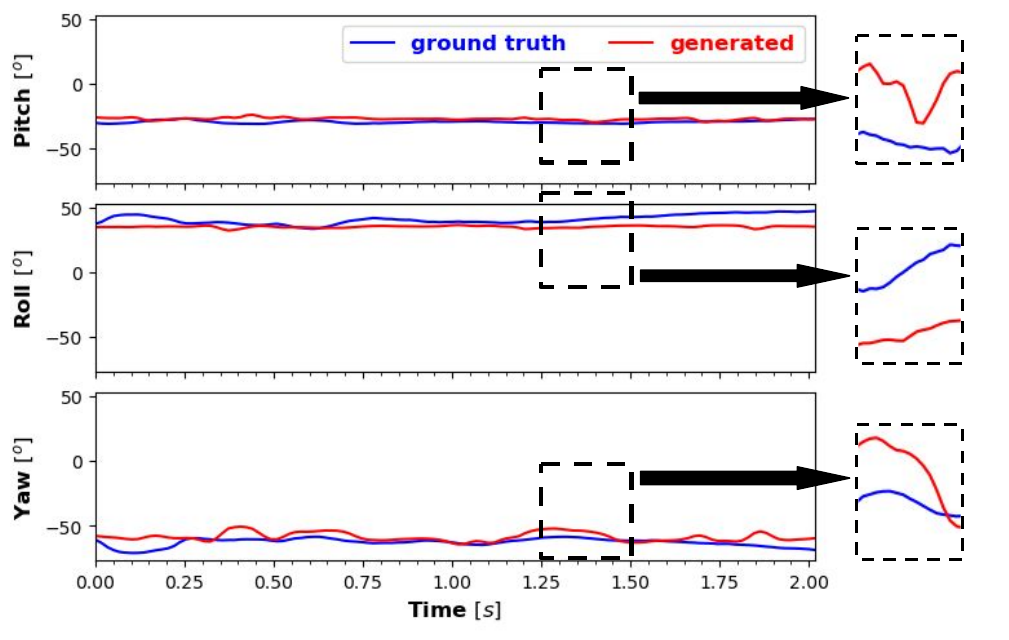}
    \caption{Sample results of generated head motion response (in terms of roll, pitch and yaw) using the proposed model with wav2vec2. Overall, the generated results closely approximate the ground truth.}
    \vspace{-3mm}
    \label{fig:plots}
\end{figure}
\vspace{-1mm}
\subsection{Results and Analysis}
\vspace{-1mm}
\textbf{Generalized head motion generation:} Table \ref{tab:results_main} compares the performance of our proposed model (in terms of MAE) with two baselines (see section 3.3) using various speech features as inputs. We note that the prediction performances for all head angles are comparable. Wav2vec2 yields the best performance for our model, while eGEMAPS works the best for the LSTM baseline. In all cases, our model can generate head motion significantly above the real time speed (of course, no rendering involved at this moment) with reasonable accuracy. The high standard deviation is attributed to our subject-independent model, which does not fine tune to specific subject's head motion. It is well known that head gesture is subjective, cultural and has individual idiosyncrasies. Fig. \ref{fig:plots} shows a visual comparison of our generated results with actual ground truth.
\par We also perform \textbf{ablation studies} to observe the contributions of learnable smoothing and cosine similarity. When wav2vec2 is used as features, cosine similarity does not make much difference for our model, although this helps in other cases. Learnable smoothing also makes a positive contribution to the model's performance except for eGEMAPS.\\

\vspace{-2mm}
\noindent\textbf{Personalised vs. generalised:} A major advantage of our approach over existing work is that it performs accurately despite being subject-independent i.e.,the training subjects and test subjects have no overlap. To compare if our model will perform even better, if personalised, we carried our subject-specific training by including data from specific listener's in the training set. Results in Table \ref{tab:dep}, shows that performance is slightly better (particularly for yaw) than the proposed generalised model. Note that the subjects in the dataset are professional actors who come from similar cultural background. Therefore, personalization does not translate to a large improvement. However, personalizing a model also limits its scope of applicability in real world scenario, as individual's data may not be available beforehand for training.\\
\begin{table}[tb]
\caption{Generalised and personalised head motion generation results using our graph-based model with Wav2vec2.}
\vspace{-2mm}
\label{tab:dep}
\centering
\small
\begin{tabular}{l|c c c}
\toprule
 & {\bf Roll} & {\bf Pitch} & {\bf Yaw} \\
 \midrule
Generalized & $3.30\pm3.57$ & $3.53\pm 2.50$ & $5.87\pm5.69$ \\
Personalised & $2.83\pm3.35$ & $2.62\pm2.33$ &$3.53\pm5.03$ \\
\midrule \midrule
{Generalized} &\multirow{2}{*}{$3.12\pm3.58$} & \multirow{2}{*}{$3.35\pm2.67$} & \multirow{2}{*}{$6.39\pm5.48$} \\
{w/ speaker's affect}  & &  &  \\
\bottomrule
\end{tabular}
\vspace{-2mm}
\end{table}

\vspace{-2mm}
\noindent\textbf{Real-time performance and model size:}
To be able to use the generated results in real world the generation of head pose angles should be faster than 30 frames per second (fps), and for the listener to be considered `human-like' the response lag should be less than 250ms \cite{realtime}. Based on the results of generation speed (includes both feature extraction and model inference) shown in Table \ref{tab:results_main}, the proposed model performs better than the required real-time speed.\\

\vspace{-2mm}
\noindent\textbf{Does speaker emotion impact listener's head motion response?} Individuals are known to align their emotional states during a conversation \cite{emotion_alignment}. To test this hypothesis, we used the arousal and valence annotations of the speaker available in the dataset as additional features in our model. Table \ref{tab:dep} shows no clear advantage when considering speaker's emotion as the results are comparable to our emotion-unaware generalised model.

%
\vspace{-1mm}
\section{Conclusions}
\label{sec:majhead}
\vspace{-1mm}
We propose an end-to-end generalised model to predict continuous head motion response of a listener using only speaker's speech during an in-person dyadic interaction. Our graph-based model generates head pose in terms of roll, pitch and yaw with an average error of 4.5 degrees. To the best of our knowledge, this is the first work on generating continuous head motion response of a listener in real time using speaker's speech alone. The relatively lower error and high generation speed of our model make it suitable for deployment in real-world human-robot interaction scenarios. However, a subjective evaluation is needed to ensure suitability. A limitation of this work is that it does not account for the context or the lexical content of speech. Our model could be improved by taking language into account, although it may slow down the system. Also, our current dataset has only ten subjects (professional actors) all from similar cultural background, while head motion is known to change with language and culture. Future work will address the dataset limitations by considering larger datasets with more and diverse subjects.
\vspace{-4mm}
\section{Acknowledgement}
\vspace{-1mm}
We thankfully acknowledge support from Intel Corporation and EPSRC DTP EP/W524359/1 for this work.

\balance
\bibliographystyle{IEEEbib}
\bibliography{strings,refs}

\end{document}